\documentclass[11pt,letterpaper]{article}

\usepackage{hyperref}
\hypersetup{
    colorlinks=true,
    linkcolor=blue,
    filecolor=magenta,      
    urlcolor=cyan,
}

\newcommand\nnfootnote[1]{%
    \begin{NoHyper}
    \renewcommand\thefootnote{}\footnote{#1}%
    \addtocounter{footnote}{-1}%
    \end{NoHyper}
}

\usepackage{fullpage}
\usepackage{parskip}
\usepackage{cite}
\usepackage{amsmath,amssymb,amsfonts}
\usepackage{mathtools}
\usepackage{textcomp}
\usepackage{algorithm}
\usepackage[noend]{algpseudocode}
\usepackage[noabbrev]{cleveref}
\usepackage{titlesec}
\usepackage{bm}
\usepackage{graphicx}
\usepackage{color}
\usepackage{xcolor}
\usepackage{colortbl}
\usepackage{float}
\usepackage{bbm}
\usepackage{comment}
\usepackage{tikz}
\usepackage{comment}
\usepackage[shortlabels]{enumitem}
\usepackage{caption}
\usepackage{subcaption}
\usepackage{minitoc}
\usepackage{stackengine}
\usepackage{times}
\usepackage{authblk}
\usepackage{acronym}

\DeclarePairedDelimiter\ceil{\lceil}{\rceil}
\DeclarePairedDelimiter\floor{\lfloor}{\rfloor}

\newacro{ATSC}{Adaptive Traffic Signal Control}
\newacro{RL}{Reinforcement Learning}
\newacro{DQN}{Deep Q-Networks}
\newacro{PER}{Prioritized Experience Replay}
\newacro{ML}{Machine Learning}
\newacro{MDP}{Markov Decision Process}
\newacro{DDQN}{Double DQN}
\newacro{TD}{Temporal Difference}
\newacro{SUMO}{Simulation of Urban MObility}
\newacro{ReLU}{Rectified Linear Unit}
\newacro{FT}{Fixed-Time}
\newacro{SOTL}{Self-Organizing Traffic Light}
\newacro{CNN}{Convolutional Neural Networks}
\newacro{GAN}{Generative Adversarial Networks}
\newacro{NN}{Neural Networks}
\newacro{IS}{Importance Sampling}

\setstackEOL{\\}
\makeatletter
\def\BState{\State\hskip-\ALG@thistlm}
\makeatother

\allowdisplaybreaks

\def\BibTeX{{\rm B\kern-.05em{\sc i\kern-.025em b}\kern-.08em
    T\kern-.1667em\lower.7ex\hbox{E}\kern-.125emX}}

\makeatletter
\def\blfootnote{\xdef\@thefnmark{}\@footnotetext}
\makeatother

\title{Adaptive Traffic Control with Deep Reinforcement Learning: Towards State-of-the-art and Beyond}

\author[1]{Siavash Alemzadeh}
\author[2]{Ramin Moslemi}
\author[2]{Ratnesh Sharma}
\author[1]{Mehran Mesbahi}

\affil[1]{William E. Boeing Department of Aeronautics and Astronautics, University of Washington, Seattle, WA, USA}
\affil[2]{NEC Laboratories America Inc., San Jose, CA, USA}

\date{}

\begin{document}

\maketitle

\nnfootnote{$*$The research was conducted when the first author was an intern at NEC Laboratories America Inc. Corresponding author email: \texttt{alems@uw.edu}.}

\begin{abstract}
    In this work, we study adaptive data-guided traffic planning and control using \ac{RL}.
	We shift from the plain use of classic methods towards state-of-the-art in deep \ac{RL} community.
	We embed several recent techniques in our algorithm that improve the original \ac{DQN} for discrete control and discuss the traffic-related interpretations that follow.
	We propose a novel \ac{DQN}-based algorithm for Traffic Control (called TC-DQN$^+$) as a tool for fast and more reliable traffic decision-making.
	We introduce a new form of reward function which is further discussed using illustrative examples with comparisons to traditional traffic control methods.
\end{abstract}


\section{Introduction}
\label{sec:Intro}

    Transportation and urban mobility are the backbones of every major city's infrastructure with direct impacts on health, environmental costs, and businesses.
	For instance, only in 2018 Americans lost 97 hours on average due to congestion, costing them nearly \$87 billion---almost \$1,348 per driver \cite{INRIX} and the demand is still steadily getting higher.
	As a consequence, there has been decades of attempts to integrate optimization and computational proficiency into traffic management and control \cite{wei2019survey}.
	Besides, with the advent of an unprecedented computing power, the research has lately observed remarkable increase in leveraging optimization and machine learning methods.

	While early optimization tools for traffic signal control (e.g. SCOOT \cite{hunt1981scoot}, SCATS \cite{lowrie1990scats}) are still employed in several large cities worldwide, they are non-adaptive in the sense that manual design for signal plans are required in practice.
	\ac{RL} methods, on the other hand, possess an adaptive characteristic in sequential decision-making of traffic signals and have shown promising progress comparing to classic strategies in transportation.
	Nevertheless, the method is yet to be officially exercised in real scenarios due to fundamental shortcomings.
	For instance, training machine learning algorithms on real-world examples is dangerous, expensive, and time-consuming; thereby, \ac{RL} models are mainly trained in simulations that often lack physical precision.
	Moreover, unlike domains with limited possible states (say, Atari games), adopting classic methods for highly complex environments such as transportation can be too ambitious.

\begin{figure}[t]
		\centering
		\includegraphics[width=0.6\columnwidth]{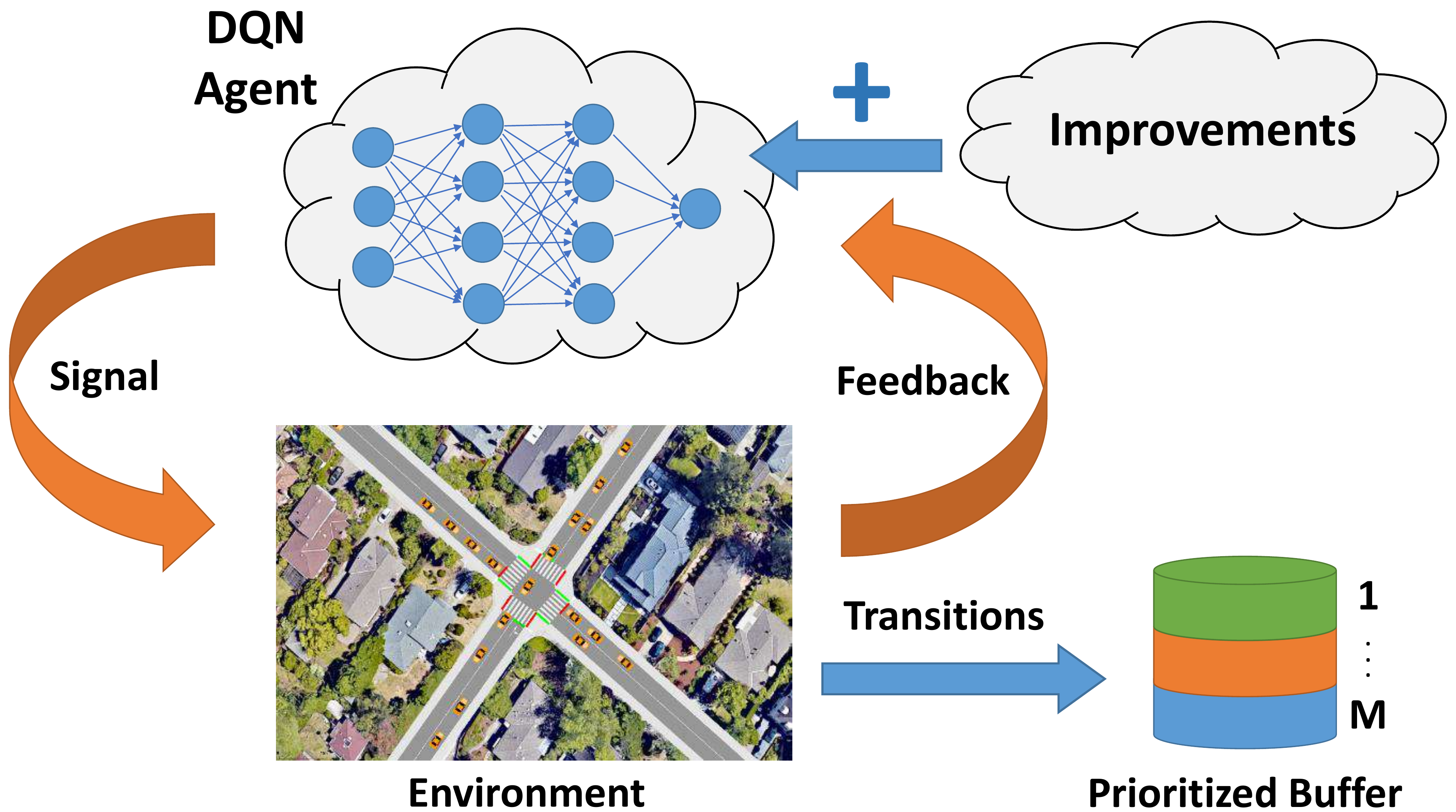}
		\caption{Schematic of our proposed algorithm.}
		\label{fig:RLscheme}
	\end{figure}

	In this work, we focus on such issues by (1) equipping our simulation environment with traffic scenarios that are created using real-world open-source data, and (2) incorporating the cutting-edge techniques in deep \ac{RL} to improve our algorithm's performance.
	In particular, inspired by the \textit{rainbow} algorithm \cite{hessel2018rainbow}, we employ latest advancements on \ac{DQN} including \textit{Double $Q$-Learning}, \textit{Dueling Networks}, and \textit{\ac{PER}}.
	We also take advantage of recent studies on novel exploration methods in \ac{RL} as well as distributional forms of reward values and showcase the effectiveness of each technique on multiple traffic scenarios.
	Furthermore, we address a new form of discretized reward function as an additional requirement due to distributional \ac{RL} technique (which is sensitive to rarely occurring events such as very smooth or highly congested traffic).
	Our empirical observations suggest that adding customized heuristics in the reward function can positively affect the performance.
	Finally, we provide illustrative case-studies wherein the benefits of our method as well as comparisons with some traditional architectures in \ac{ATSC} are demonstrated on simulations that follow real traffic scenarios.
	Schematic of the idea behind our method is given in \Cref{fig:RLscheme}.
	
	The rest of the paper is organized as follows: \Cref{sec:relWork} reviews the related literature.
	In \Cref{sec:background}, we provide some background.
	\Cref{sec:probForm} defines the problem formulation followed by the experimental results in \Cref{sec:experiments}.
	Conclusions and future directions are addressed in \Cref{sec:conclusion}.
    
\section{RELATED WORK}
	\label{sec:relWork}
	
	Historically, traffic control has been one of the on-the-spot applications in science and engineering heavily due to the huge incurred expenses, health issues, and depleted natural resources that are caused from urban congestion and pollution.
	Moreover, research in traffic planning is highly restrictive in the sense that a newly proposed methodology should comprise: (1) computationally efficient implementation, and (2) minimal impractical assumptions.
	This line of work dates back to 1980's with the advent of offline traffic light timing methods such as MAXBAND \cite{little1981maxband} where the scheme is selected through a self-regulatory process of multi-timing control which, in turn, was further improved using dynamic adjustment of parameters in methods such as SCOOT \cite{hunt1981scoot} and SCATS \cite{lowrie1990scats}.
	However, these products are suitable for regional traffic control and in many cases require considerable human intervention.
	Next generations of dynamic traffic control emerged in the following years (e.g. RHODES \cite{mirchandani2001real}) to address online control of time-varying traffic flows.
	This made the scheme more realistic but, meanwhile, highly expensive computationally.
	In general, the aforementioned methods face fundamental shortcomings including simplification of traffic constraints and lack of timely response to the actual traffic fluctuation \cite{wei2019survey, wang2018review}

	Fast forward a few decades, the so-called big data revolution has led to the integration of ideas from data-driven control and machine learning into transportation research.
	In particular, \ac{RL} has been widely used to address traffic control due to its model-free nature \cite{mikami1994genetic, prashanth2010reinforcement}.
	The results became even more promising after the breakthrough in deep \ac{RL}, \ac{DQN} \cite{mnih2015human}.
	Thereafter, the \ac{ATSC} literature has been heavily focused on what type of \ac{RL} algorithm to pick\footnote{Namely, \textit{model-based} vs. \textit{model-free} \ac{RL} or \textit{policy-based} vs. \textit{value-based} learning.} \cite{genders2019open} and how to define \ac{RL} parameters---i.e. states, rewards, and actions---to further enhance those algorithms \cite{wei2018intellilight}.
	Nevertheless, more recent advancements in \ac{RL} are less tackled in traffic control literature.
	It is shown that the overall combination of these techniques results in a significantly improved performance on several instances of Atari games \cite{hessel2018rainbow}.
	Similarly, we show that such combination plus heuristics in the \ac{RL} problem definition leads to a substantially improved convergence on traffic control scenarios.
	A list of such methods includes---but is not limited to--- \emph{Double Q-Learning} \cite{van2016deep},  \emph{Prioritized Replay} \cite{Schaul2016PrioritizedER}, \emph{Dueling Networks} \cite{wang2016dueling}, \emph{NoisyNets} \cite{fortunato2018noisy}, and \emph{distributional \ac{RL}} \cite{bellemare2017distributional}.
	TC-DQN$^+$ is an endeavour to show how such combinations contribute to the particular case of \ac{ATSC}.


\section{BACKGROUND}
	\label{sec:background}
	
	\subsection{RL Notations and Background}
	\label{subsec:RL}
	
	An \ac{RL} problem is defined as a \ac{MDP} tuple $<\mathcal{S}, \mathcal{A}, \mathcal{T}, r, \gamma>$ where the parameters respectively denote the set of states, set of possible actions, transition function, reward, and discount factor.
	At time-step $t$, the environment provides the agent with an observation (state) $s_t\in\mathcal{S}$, and the agent selects some action $a_t\in\mathcal{A}$.
	This interplay induces the transition $\mathcal{T}:\mathcal{S}\times\mathcal{A}\rightarrow\mathcal{S}$ defined as $\mathcal{T}(s,a,s')=\mathbb{P}(s_{t+1}=s'|s_t=s,a_t=a)$.
	As a consequence of taking the action, the environment provides the corresponding reward $r:\mathcal{S}\times\mathcal{A}\rightarrow\mathbb{R}$ defined at the state-action pair $(s,a)$ as $r_t(s, a)=\mathbb{E}[\mathcal{R}_t|s_t=s, a_t=a]$ and also the resulting state $s_{t+1}$ where $\mathcal{R}_t$ denotes the one-step reward.
	We assume an episodic \ac{MDP} with constant $\gamma\in(0,1]$ and the goal of the agent is to maximize the expected return  $G_t = \sum_{k=0}^{\infty}\gamma^k\mathcal{R}_{t+k}$ by finding an optimal policy $\pi^*$ through an iterative trial-and-error process.
	The \ac{RL} agent learns the \emph{state-action} value function ($Q$-function) $Q^{\pi}(s,a) = \mathbb{E}_{\pi}[G_t|s_t=s, a_t=a]$ following policy $\pi$ after action $a$ is taken at time $t$.
	The optimal value $Q^*(s,a)=\max_{\pi}Q^\pi(s,a)$ satisfies the Bellman optimality equation for $s\in\mathcal{S}$ and $a\in\mathcal{A}$,
	\begin{align}
		\label{eq:q-bellman}
		Q^*(s,a) = \sum_{s'\in\mathcal{S}} \mathcal{T}(s,a,s') [r_t(s,a) + \gamma\max_{a'} Q^*(s',a')].
	\end{align}

	\subsection{Deep Q-Networks and Improvements}
	\label{subsec:DQN}
	
	Unlike classic \ac{RL} methods, \ac{DQN} employs deep \ac{NN} to approximate the $Q$-function in a nonlinear fashion, making \ac{RL} applicable to high-dimensional problems \cite{mnih2015human}.
	In addition, the work uses two other pivotal techniques: using an \emph{experience replay buffer} to remove data correlation and also integrating a less frequently updated \emph{target network} to prevent fluctuations of $Q$-values in the \ac{TD} term of the loss defined as,
	\begin{equation}
		\begin{aligned}
			\label{eq:DQNLoss}
			\mathcal{L}(\theta) = \mathbb{E}_{s,a,r,s'\sim\mathcal{D}} \Big[ \Big( \text{TD}_{\textit{target}} - Q(s,a;\theta) \Big)^2 \Big],
		\end{aligned}
	\end{equation}
	where $\theta$ contains the parameters of the \ac{NN} and $\text{TD}_{\textit{target}}=r_t(s,a) + \gamma\max_{a'} Q(s',a';\theta^-)$ where $\theta^-$ denotes periodic copies of $\theta$ and $\mathcal{D}$ is the (potentially time-varying) underlying distribution.
	Despite the initial success, the original \ac{DQN} algorithm lacks stable and fast convergence in many cases due to poor exploration ($\epsilon$-greedy) and \ac{NN} structure.
	There has been many independent attempts to improve the performance of \ac{DQN} over the past few years.
	For instance, \emph{Double $Q$-Learning} \cite{van2016deep} addresses the overestimation bias in the maximization step in \eqref{eq:q-bellman} resulting in the modified $\text{TD}_{\textit{target}}$,
	\begin{equation}
		\begin{aligned}
			\label{eq:new_TD}
			\text{TD}_{\textit{target}} = r_t(s,a) + \gamma Q(s',\text{argmax}_{a'} Q(s',a';\theta);\theta^-).
		\end{aligned}
	\end{equation}
	Regarding the Experience Replay, \emph{\ac{PER}} \cite{Schaul2016PrioritizedER} is an effort to assign higher priorities to those \ac{RL} transitions with higher \ac{TD}-error.
	New data are then inserted into a replay buffer with maximum priority, providing a bias towards more recent data.
	Another shortcoming of \ac{RL} is that learning the expectation of returns, $G_t$, can be restrictive as the multimodality and the variance of similar actions (values) may not be preserved.
	This has been investigated over the years \cite{morimura2010parametric}, and subsequently led to the idea of learning distributions---rather than expectation---of returns, referred to as \emph{distributional \ac{RL}} \cite{bellemare2017distributional}.
	Lastly, there has been improvements due to \ac{NN} architecture modifications.
	Namely, the \emph{Dueling} Network structure \cite{wang2016dueling} separates the learning streams of the value function and the advantage function.
	This lends more flexibility to learning in the sense that redundant actions become less influential when the value function prevails at some specific state.
	Furthermore, \emph{NoisyNets} \cite{fortunato2018noisy} modify the \ac{NN} structure by adding noise to some hidden layers, whereas, unlike $\epsilon$-greedy, the parameters of the noise are adjusted by the model itself during training.
	This allows the agent to decide when and in what proportion to append uncertainty in \ac{NN} parameters, making the exploration more efficient.
	\noindent In general, the methods introduced in this section have entailed promising improvement over the original \ac{DQN} in accelerating the training and convergence.
	We will modify and utilize these techniques in \ac{ATSC}.


\section{PROBLEM FORMULATION}
	\label{sec:probForm}
	
	\subsection{RL Environment and Agent}
	\label{subsec:RL_env_and_agent}
	
\begin{figure}[t]
		\centering
		\includegraphics[width=0.4\columnwidth]{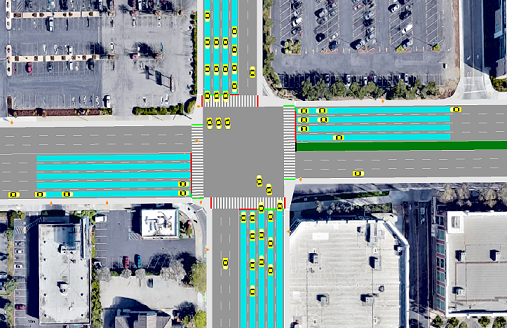}
		\caption{Snapshot of the \ac{RL} environment. The blue strips show detectable range of sensors.}
		\label{fig:env}
	\end{figure}

	The \ac{RL} environment is represented by a given intersection including vehicles with predefined probabilistic commutes (\Cref{fig:env}).
	Containing the entire traffic regulation, the environment acts as a framework to provide feedback to the \ac{RL} agent.
	We train our algorithm on \ac{SUMO}---an open-source and portable traffic simulator \cite{krajzewicz2002sumo}.
	Traffic scenarios are generated in \ac{SUMO} and adjusted from real-world data and observations (see \Cref{sec:experiments}).
	For brevity, in what follows the letters $N,E,S,$ and $W$ denote North, East, South, and West respectively, and $X \rightarrow Y$ implies from $X$ towards $Y$ ($X \leftrightarrow Y$ denotes both directions).

	The \ac{RL} agent is the decision-maker for the traffic light whose actions dictate the signal phase and cycle duration.
	In response, the agent receives a reward and the environment is shifted to the next iteration.
	The proper choice of \ac{RL} \emph{states}, \emph{actions}, and \emph{rewards} is crucial to any \ac{RL} algorithm.
	We describe our selection as well as traffic-related interpretations of these parameters.

\begin{figure}[t]
    	\centering
    	\includegraphics[width=0.5\columnwidth]{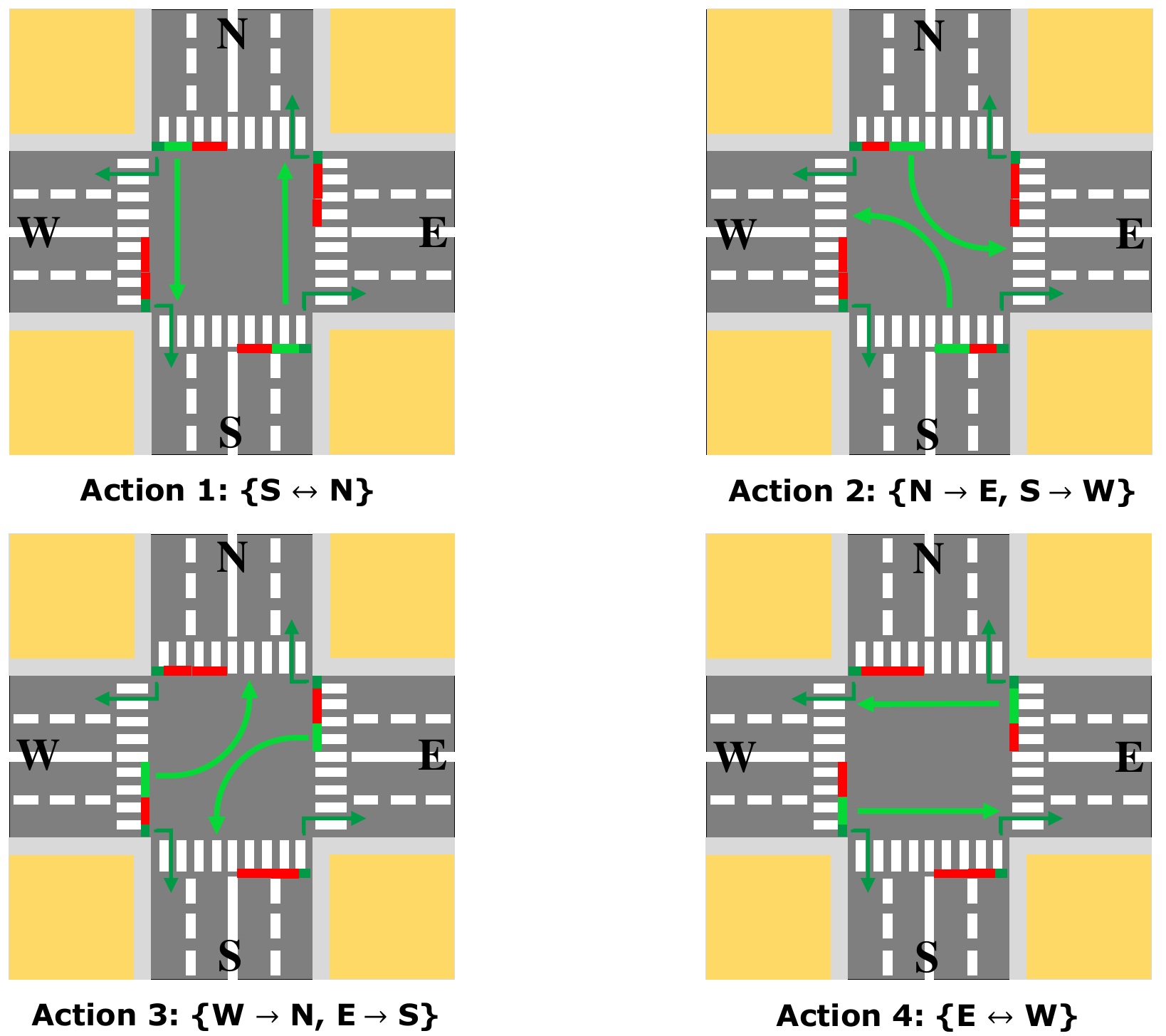}
    	\caption{An example of the action set for a given intersection.	Dark green implies yield to crossing lane.}
    	\label{fig:actions}
    \end{figure}

	\textbf{\textit{Actions.}} The set of actions, $\mathcal{A}$, includes available traffic signal patterns (\Cref{fig:actions}).
	The signal pattern is assumed to be \emph{non-sequential}, i.e., the agent is free to choose any pattern (action) $a\in\mathcal{A}$ at each time-step.
	We set a minimum time, $T_g$, for each green light period to avoid sporadic flickers of the traffic light.
	The agent is free to stick to the current phase or move on to a fixed yellow phase, $T_y$, followed by an all-red interval, $T_r$, that are predefined to maintain smooth traffic.\footnote{There is a trade-off in the choice of $T_g$, where lower values encourage more flexibility, whereas high values result in more learning stability.}
	If the agent holds on to its current action, then $T_y=T_r=0$.
	Overall, the performance of our algorithm is highly sensitive to the integers $T_g, T_y,$ and $T_r$.

	\textbf{\textit{States.}} The state at time $t$, $s_t\in\mathcal{S}$, is a real vector that contains the information received from the environment.
	Often a number of measured elements (based on available sensing technology) are integrated to form the state vector (such as total wait time, queue length, etc. \cite{aslani2017adaptive}).
	However, the usefulness of excessive application of various	measures for the state is still questionable \cite{zheng2019diagnosing}.
	In our analysis, we set the state as the total number of vehicles within distance $d$ of the intersection on each lane (\Cref{fig:env}) and collect those integers for $T_g$ consecutive frames after each action is taken.
	\noindent The aggregation of these frames and a one-hot key element\footnote{We use one-hot encoding with a slight abuse of notation and choose a scalar (instead of a vector) since the action space is not large in our setup.}---that determines the current phase---are concatenated to form $s_t\in\mathbb{N}_{\{0\}}^n$ where $n = |\mathcal{A}| T_g + 1$ and $|\mathcal{A}|$ denotes the cardinality of $\mathcal{A}$.
	Note that such definition implicitly captures vehicles' speeds without requiring to add extra dimensions, resulting in lower complexity and faster learning.

	\textbf{\textit{Rewards.}} Reward function, $r_t$, is a pivotal part of \ac{RL} algorithm design as it is the feedback index to evaluate the quality of the agent's performance.
	Common reward functions in traffic control are defined as a combination of elements such as queue length, wait time, throughput, etc., mostly accompanied with incidental factors (frequency of signal change, accident avoidance, etc.) \cite{nishi2018traffic, arel2010reinforcement}.
	However, defining the reward function in this way is often not directly aimed at minimizing the total travel time of the vehicles at the intersection.\footnote{Since the optimization step occurs immediately after each action and the episodic sequential effects are neglected in the learning process.}
	Moreover, such rewards are highly sensitive to the weight of each factor, which makes the agent incapable of distinguishing between different learning measures \cite{wei2018intellilight}.

	We take a novel approach and define the reward function for state-action pair $(s,a)$ as,
	\begin{equation}
		\begin{aligned}
			r(s,a) =  - \Big( r_a(s) + \mathbbm{1}_{\{t=T\}} r_e \Big).
		\end{aligned}
	\label{eq:reward}
	\end{equation}
	
	The \emph{action-based reward}, $r_a$, captures the reward corresponding to action $a$ for $T_g$ frames and is defined as,
	\begin{equation}
		\begin{aligned}
			r_a(s) = \sum_{t=t_a}^{t_a+T_g} \Bigg( p_1 \sum_{k=1}^{N_a} \omega_t^k(s) + p_2 \mathbbm{1}_{\{N_t=0\}} \Bigg) + p_3\mathbbm{1}_{\{ t_a \neq t_{a'} \}},
		\end{aligned}
		\label{eq:ra}
	\end{equation}
	where $\omega_t^k$ is a binary variable indicating whether vehicle $k$ is waiting at the intersection at time $t$, $t_a$ is the time-step at which action $a$ is employed, and $N_a$ is the total number of vehicles that pass through the intersection in the interval of $[t_a,t_a+T_g]$.
	The term $\mathbbm{1}_{\{N_t=0\}}$ is considered to specify a penalty in case of a phase shift when there is no vehicle waiting---discouraging redundant actions.
	Further, the term $\mathbbm{1}_{\{ t_a \neq t_{a'} \}}$ indicates whether two consecutive signal phases are equal and is considered to oppose sporadic flickers of the traffic light.
	The variables $p_1$, $p_2$, and $p_3$ are the penalty weights.
	Overall, defining action-based rewards (as in Eq. \eqref{eq:ra}) is advantageous since it (1) enables the agent to have a measure of what actions lead to higher rewards, (2) naturally accounts for the temporal aspects of information such as "speed of vehicles", and (3) avoids unnecessary actions.

\begin{figure}[t]
		\centering
		\includegraphics[width=0.5\columnwidth]{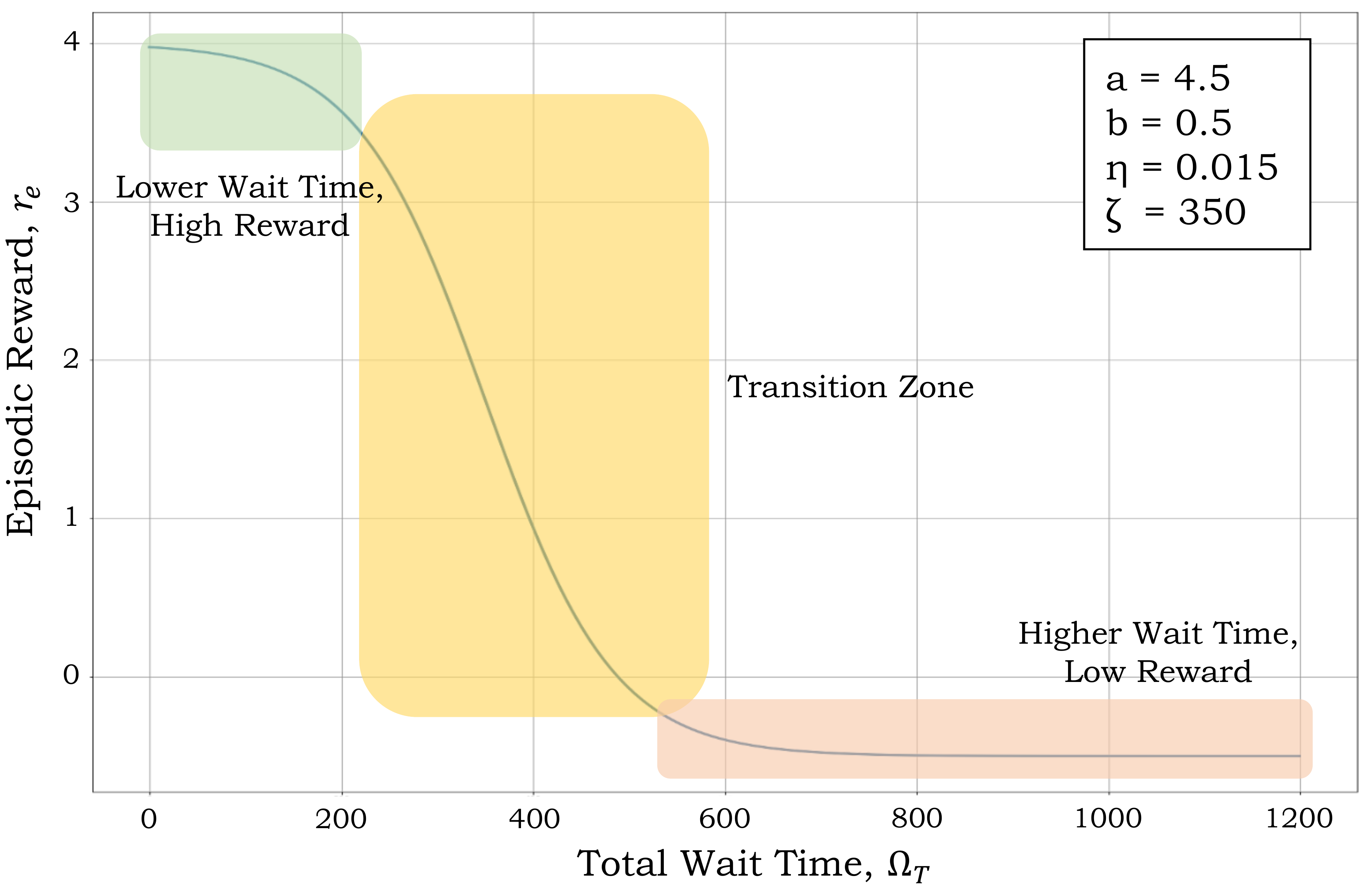}
		\caption{An example of the episodic reward function, $r_e$.}
		\label{fig:sigmoid}
	\end{figure}

	The \emph{episodic reward}, $r_e$, is a measure index to evaluate success or failure at the termination of each episode ($T$ indicates the length of an episode).
	However, in opposition to binary terminal rewards\footnote{i.e. to evaluate the performance based on a win/lose situation after the termination of an episode---for instance, in Atari games.}, we define,
	\begin{equation}
		\begin{aligned}
			r_e(\Omega_T) = a \sigma \big( \eta (\Omega_T + \zeta) \big) + b,
		\end{aligned}
		\label{eq:re}
	\end{equation}
	where $\sigma(x)=1/(1+e^{-x})$ is the Sigmoid function, $\Omega_T = \sum_{k=1}^{N_e} \sum_{t=0}^{T} \omega_{t}^k$ is the total wait time of all vehicles in the corresponding episode, $N_e$ is the total wait time of all of the vehicles passing through the intersection during the episode $e$, and the parameters $a, \eta, \zeta,$ and $b$ depend on the particular traffic scenario.
	Unlike the action-based reward, the episodic reward captures the influence of ``sequence of actions'' throughout episode $e$.
	As demonstrated in \Cref{fig:sigmoid}, the way we define $r_e$ results in a region of high rewards due to lower wait times (green) of the total number of vehicles throughout the episode.
	This is followed by a transition interval from high to low rewards (yellow).
	The steepness of this region can be adjusted with $\eta$.\footnote{The idea of a continuous episodic reward function becomes even more critical in complex environments where it is difficult to obtain specific terminal rewards \cite{sutton2018reinforcement}.}
	%
	%
	Finally, the transition induces a plateau region (red) where rewards worth similarly with respect to the wait time.
	%
	%
	
	\subsection{The Neural Networks Structure}
	\label{subsec:neuralNet}
	
\begin{figure*}[t]
		\centering
		\includegraphics[width=0.96\textwidth]{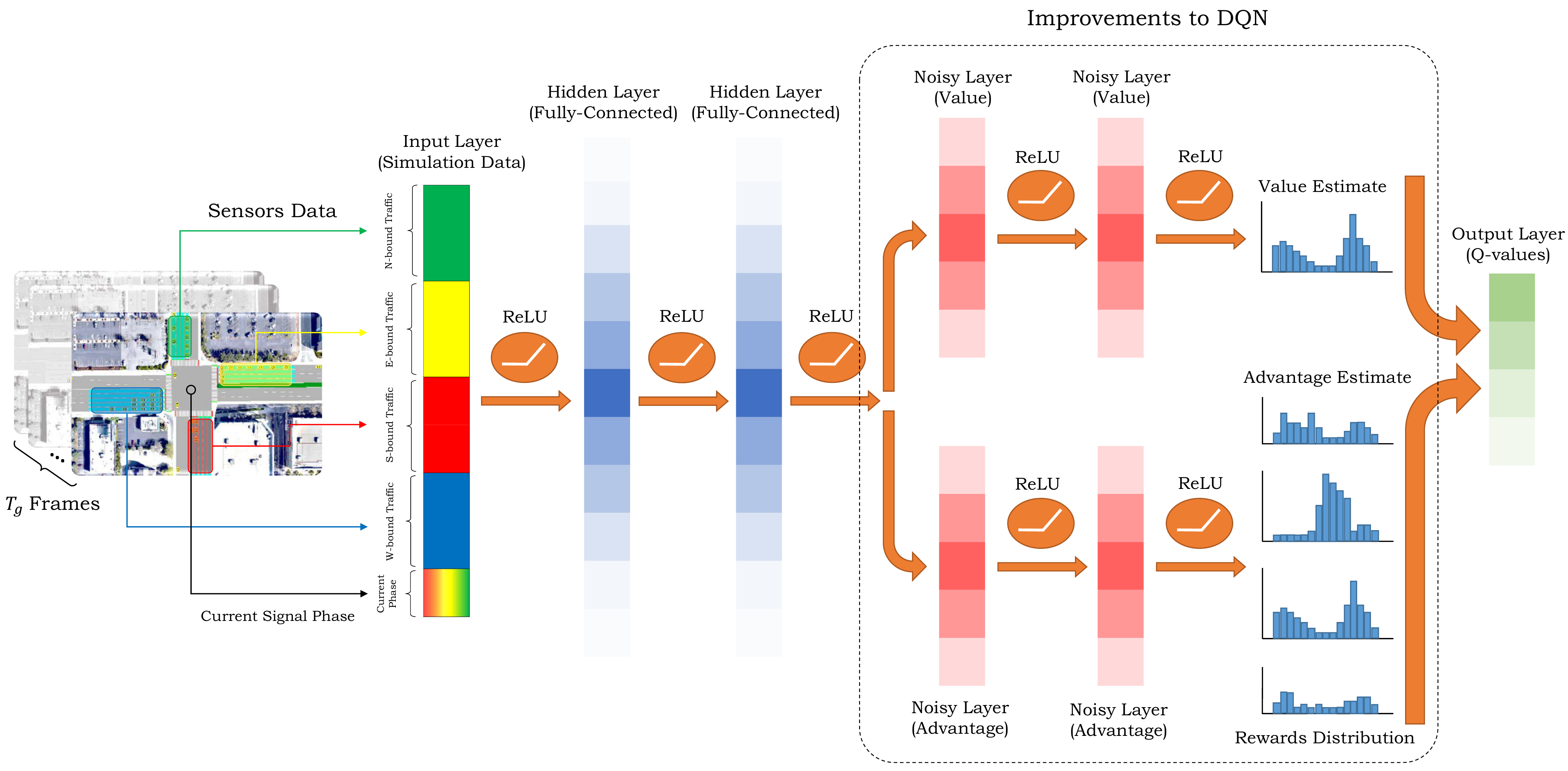}
		\caption{The structure of our proposed \ac{NN}. The dashed box contains modifications to the \ac{DQN} network.}
		\label{fig:neuralNet}
	\end{figure*}

	We use a multilayered \ac{NN} to estimate the $Q$-function.
	\Cref{fig:neuralNet} depicts a schematic of the \ac{NN} structure in our proposed algorithm.
	The input layer of the network contains data from sensors and the output layer assigns values to actions in $\mathcal{A}$.
	As discussed in \Cref{subsec:RL_env_and_agent}, the input and output layers contain $n=|\mathcal{A}| T_g + 1$ and $|\mathcal{A}|$ neurons respectively.
	%
	%
	The input layer is followed by two hidden fully-connected layers with $n_{fc}$ neurons.
	The nonlinear approximation is exploited by \ac{ReLU} for activation in our \ac{NN} setup.
	Without the dashed box in \Cref{fig:neuralNet}, original \ac{DQN} algorithm is achieved.
	We further enhance the machinery using the tools we introduced in \Cref{subsec:DQN}.
	
	In summary, we split the network into value and advantage streams, each of which accommodating two hidden noisy layers with $n_{nl}$ neurons.
	The value (upper) stream results in one discrete probability distribution, where similar distributions are assigned to each action for the advantage (lower) stream.
	The output of these two streams are summed up to the final output layer.
	The output vector is then plugged into the loss function in \eqref{eq:DQNLoss} to update the weights of \ac{NN} through backpropagation.
	As in \ac{DQN}, at each optimization step, mini-batches of data are acquired from a prioritized replay buffer with a specified memory size.
	See \Cref{appendix} for notes on reproducing the algorithm.


\section{EXPERIMENTS}
	\label{sec:experiments}
	
	\subsection{Environment Design}
	\label{subsec:envDesign}
	
\begin{figure*}[b]
		\centering
		\includegraphics[width=0.95\textwidth]{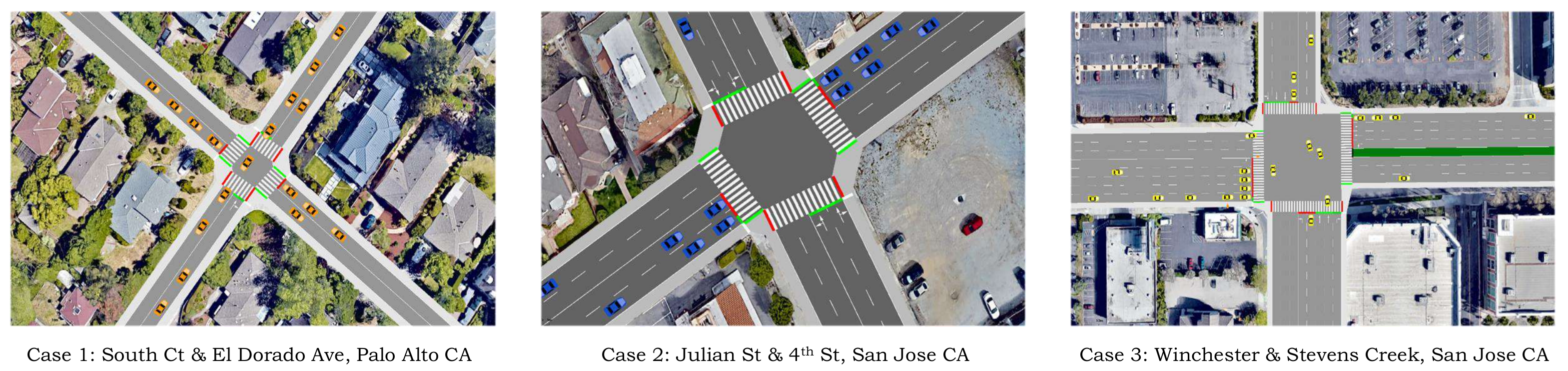}
		\caption{Case studies. The cases vary due to complexity of $\mathcal{S}$ and number of possible actions $|\mathcal{A}|$.}
		\label{fig:caseStudies}
	\end{figure*}

	In this section, we provide experiment results and compare them with baselines.
	We consider three environments that differ based on the number of actions and complexity of the state space.
	The environments are inspired by real intersections in San Francisco Bay Area (\Cref{fig:caseStudies}).\footnote{The plots from SUMO are augmented with the snapshots of the real intersection taken from Google Maps (\texttt{https://www.google.com/maps}).}
	In-person observations integrated with real data (from \cite{openData}) are given to \ac{SUMO} and the environmental hyperparameters are averaged over their real values.
	The result of the investigation is then reflected onto the vehicle generation rate as well as physical properties of the intersection such as minimum phase cycle ($T_g$, $T_y$, and $T_r$), number of lanes, right of way, etc., within the simulation environment.
	Meanwhile, to respect stochasticity of traffic flows we define,
	\begin{align}
	    X\sim\mathcal{B}(T,P_e), \hspace{4mm} \text{with} \hspace{4mm} P_e\sim\mathcal{U}(\ell,h),
	    \label{eq:flowprob}
	\end{align}
	where the random variable $X$ denotes the rate of traffic flow generated from a binomial distribution $\mathcal{B}$ with probability $P_e$ of vehicle generation per episode.
	We further assume that $P_e$ is a random variable uniformly sampled from the interval $[\ell,h]$ in order to enforce the entire range from low ($\ell$) to high ($h$) traffic flow randomly in the training stage.
	Our empirical results show that such randomness improves generalizability of our model to more traffic scenarios.

	\subsection{Results}
	\label{subsec:results}
	
	The results of our experiments are threefold.
	First, we provide training trajectories to emphasize the improvement in the convergence of the algorithm.
	Second, we test our method on three different synthetic scenarios to show the generalizability of our algorithm when facing inconsistent traffic patterns.
	Finally, we provide ablation plots to qualitatively realize the contribution of each technique (from \Cref{subsec:DQN}).
	We compare our results with the original (vanilla) \ac{DQN} in addition to two other standard classic methods: (1) \emph{\ac{FT}} \cite{miller1963settings} that utilizes a pre-specified phase cycle---mostly practiced for steady traffic---and (2) \emph{\ac{SOTL}} \cite{cools2013self} that functions when the number of waiting cars on a specific lane is above some hand-tuned threshold.
	
	\begin{figure*}[t]
		\centering
		\includegraphics[width=0.95\textwidth]{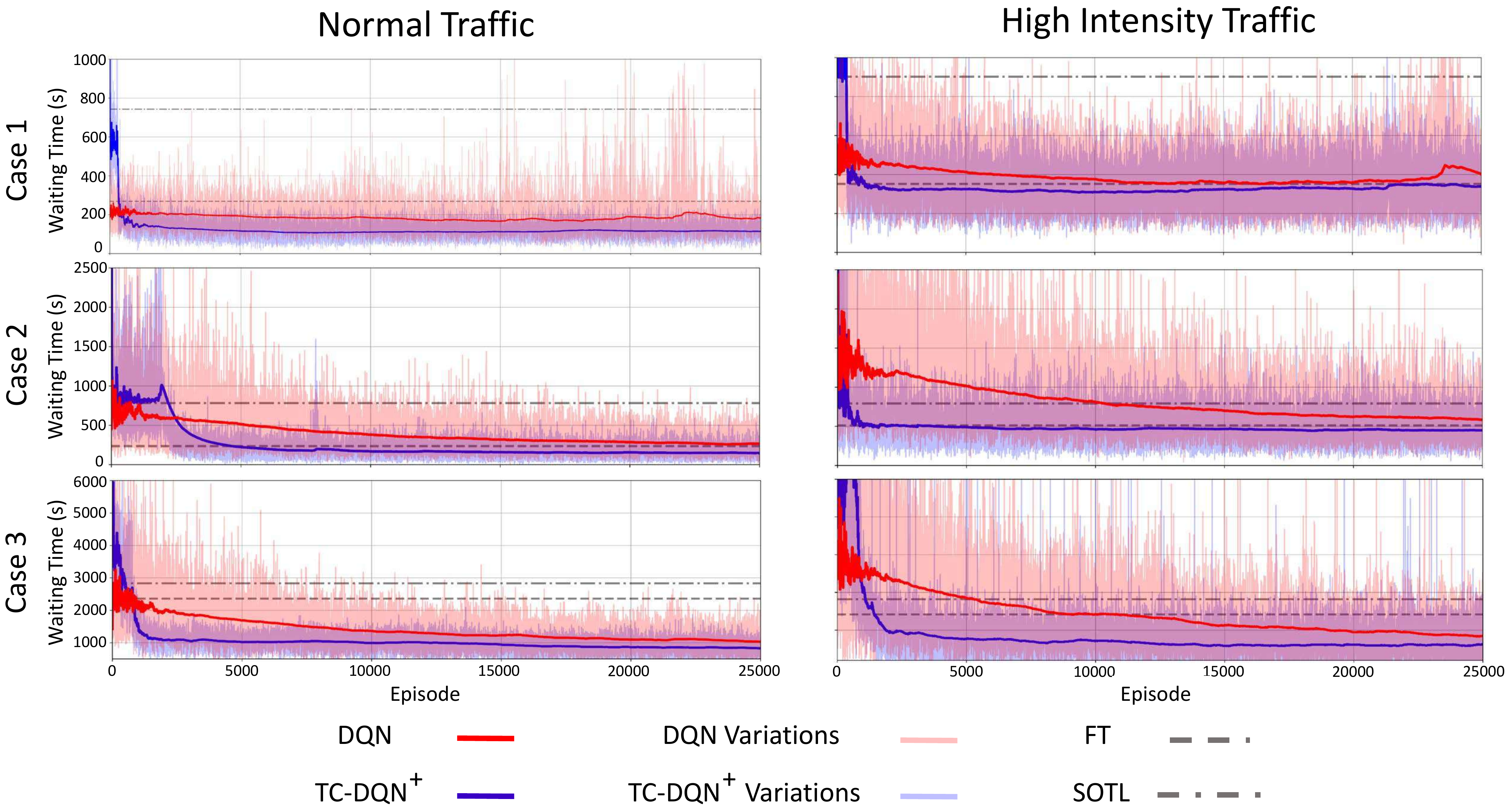}
		\caption{Total wait time learning trajectory of the vehicles at different intersections and traffic levels.}
		\label{fig:learningPlots}
	\end{figure*}
	
	
	\textit{1) Convergence Results.}
	We train our model on all three environments for two different traffic flows.
	First, we conduct training on the normal traffic patterns and then increase the flow to make the environment more complex and, hence, the learning more involved.
	\Cref{fig:learningPlots} depicts the total wait time of the vehicles based on the number of episodes the \ac{RL} agent has been trained.
	The transparent red and blue lumps show the true total wait times of {DQN} and TC-DQN$^+$ respectively at each episode and the bold curves denote the corresponding (decaying) weighted average.
	Moreover, the (average) wait time of \ac{FT} and \ac{SOTL} are also displayed for comparison.
	The monotonicity of the learning curves suggests the adaptive nature of \ac{RL} algorithms in general (that methods such as \ac{FT} and \ac{SOTL} lack).
	However, the large gap between the behavior of TC-DQN$^+$ and \ac{DQN} trajectory---specifically, in the first 10,000 thousand episodes---verifies the effectiveness of the improvements to the \ac{DQN}.
	Note the increase in the gap as the environment complexity and the traffic intensity get higher.
	In particular, \Cref{tab:timeSave} gives the time comparison between \ac{DQN} and TC-DQN$^+$.
	The entries display $\Delta\Omega_T$ for $T=400s$.
	For instance, at a relatively large intersection (case 3) and for normal traffic, an approximate of 200s total travel time is saved.
	For an average number of 30 vehicles commuting at this intersection for $T=400$ seconds, this becomes roughly 6s/vehicle of saved time.\footnote{Assuming that this could on average save 2 min/day, each vehicle saves roughly $2\times365/24\approx30$ h/year. 
	Considering the 97 hours of US traffic total wait time per driver in 2018 \cite{INRIX}, the improvement could be substantial.}
	
	\begin{table}[b]
		\centering
		\includegraphics[width=0.55\columnwidth]{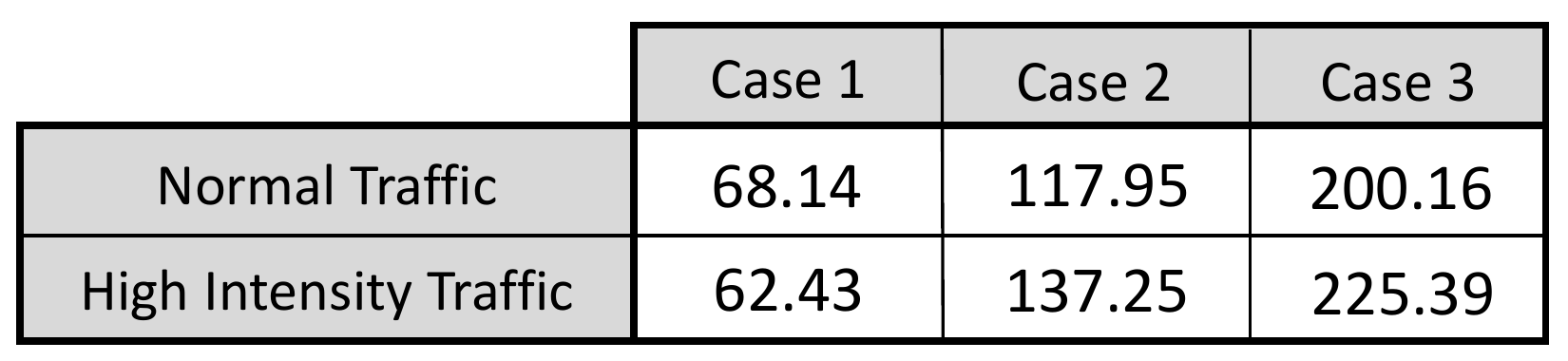}
		\caption{Time gap $\Delta\Omega_T=\Omega_T^{\text{TC-DQN}^+}-\Omega_T^{\text{\ac{DQN}}}$ (in seconds).}
		\label{tab:timeSave}
	\end{table}
	
\begin{figure}[t]
    	\centering
	    \begin{minipage}{.45\textwidth}
            \centering
            \includegraphics[width=0.85\columnwidth]{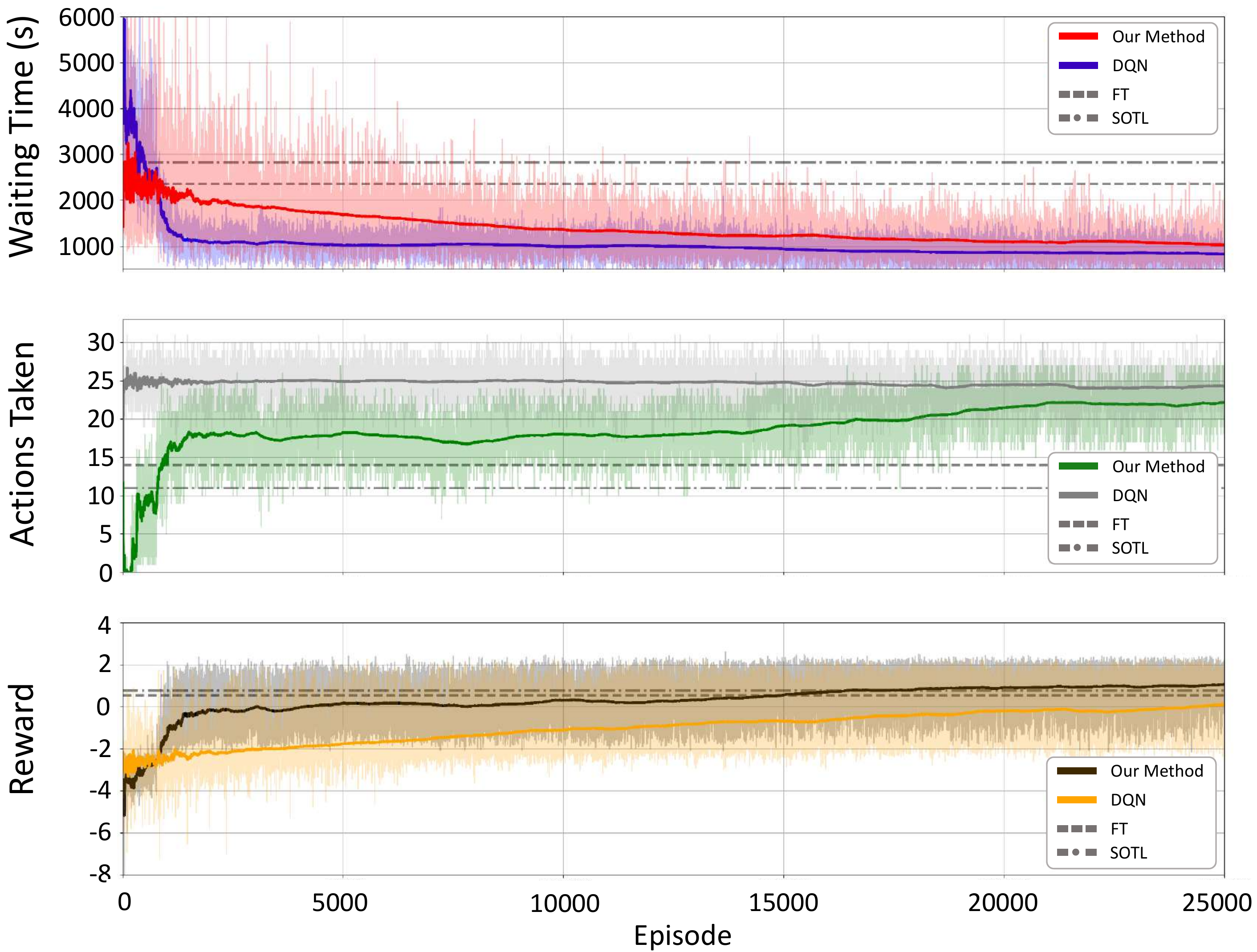}
            \caption{Training components in parallel.}
            \label{fig:learningPlots}
        \end{minipage}%
        \hspace{7mm}
        \begin{minipage}{.45\textwidth}
            \centering \includegraphics[width=1\columnwidth]{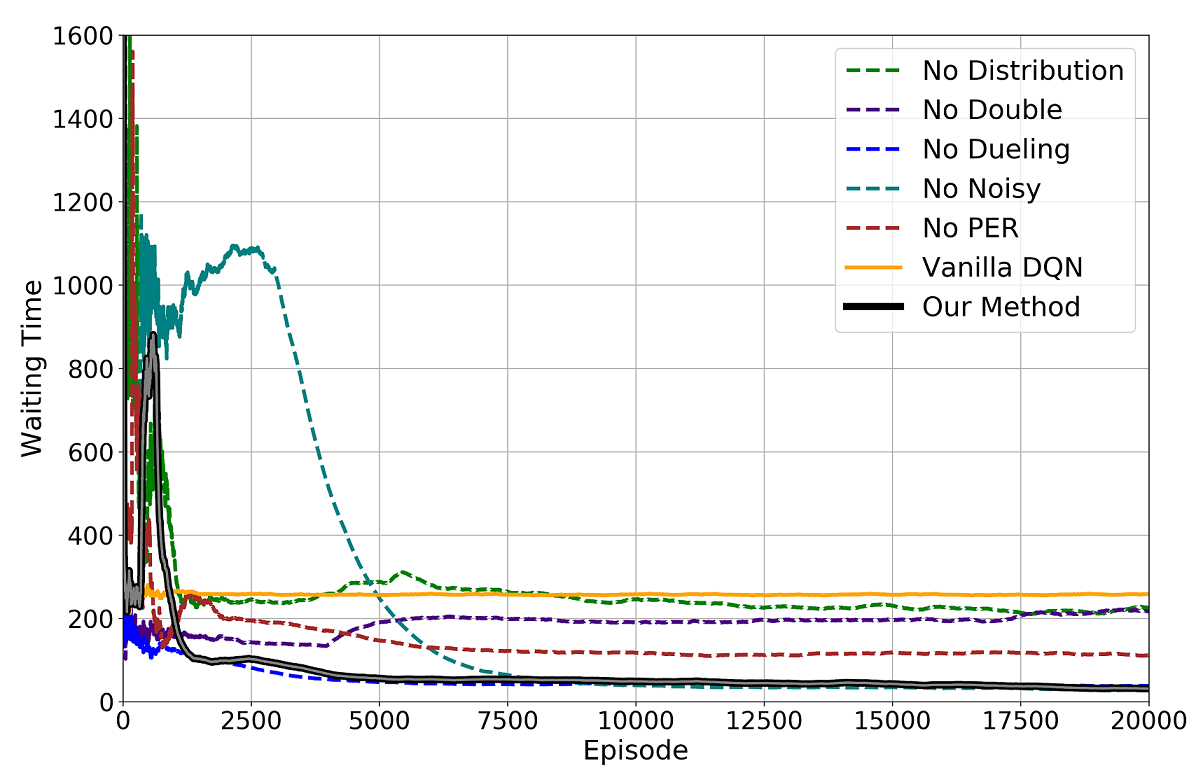}
            \caption{Ablation plots.}
            \label{fig:example}
        \end{minipage}%
    \end{figure}

	It is notable that while the standard measure of performance that we also consider is the ``total wait time'' of all vehicles ($\Omega_T$), the decrease in the variations (i.e. deviation of the true wait time from the weighted average) of \ac{RL} algorithms shall not be neglected.
	These variations help to make sense of the robustness of the algorithm when facing unseen traffic scenarios.
	As \Cref{fig:learningPlots} suggests, variations of TC-DQN$^+$ enjoy fast convergence whereas vanilla \ac{DQN} shows regular unpredictable behavior in the long-term as the $\epsilon$-greedy exploration decays over time.
	\begin{table*}[t]
		\centering
		\includegraphics[width=0.93\textwidth]{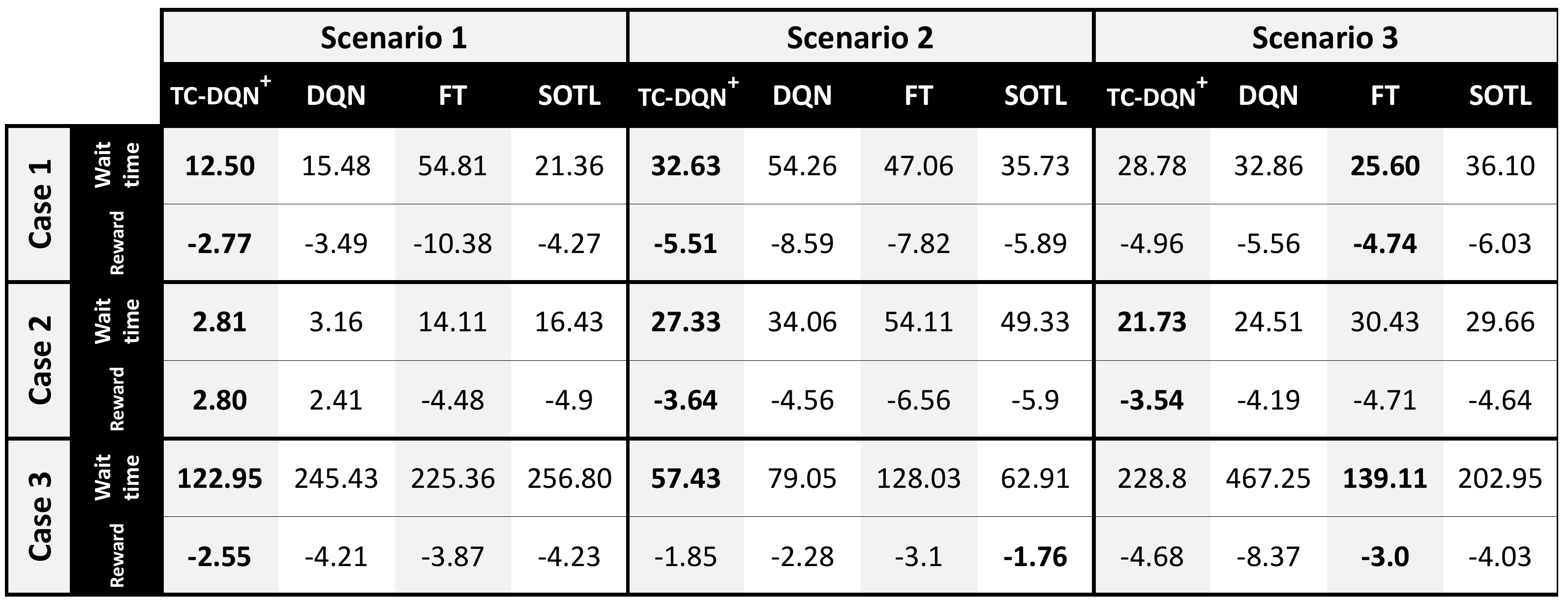}
		\caption{Results on three environments (times in minutes). Lower wait time and higher reward imply better performance.}
		\vspace{-1mm}
		\label{tab:TestCases}
	\end{table*}
	\Cref{fig:example} demonstrates the pertinence of wait time minimization to the exploration that leads to the convergence of actions and the rewards collection process in parallel (sampled for normal traffic scenario at case 3).
	The figure confirms the crucial role of efficient exploration (due to \emph{NoisyNets}) and rewards collection (due to \emph{Distributional \ac{RL}}).
	
	\textit{2) Test Cases.}	To test the performance of our model, we provide test cases on three synthetic scenarios defined as follows:
	\begin{itemize}
		\item \textbf{Scenario I.} Traffic on $E\leftrightarrow W$ is higher than $S\leftrightarrow N$.
		\item \textbf{Scenario II.} Traffic on $S\leftrightarrow N$ is higher than $E\leftrightarrow W$.
		\item \textbf{Scenario III.} Random equal traffic in all directions.
	\end{itemize}
	The results of the test cases are summarized in \Cref{tab:TestCases}.
	As the table demonstrates, TC-DQN$^+$ outperforms others in most of the traffic scenarios and environments.
	The results suggest that our algorithm performs best in traffic patterns wherein a pair of parallel directions (say, $S\leftrightarrow N$) are highly more crowded than the crossing routes ($E\leftrightarrow W$), and in some cases the improvement is substantial.
	For instance, for traffic scenario 1 at environment 3, $\Omega_T$ could be reduced by almost 2 hours in total comparing to the second best planner (\ac{FT}) when the traffic is highly concentrated on $E\leftrightarrow W$.
	Also, TC-DQN$^+$ surpasses \ac{DQN} in every case, verifying the positive contributions of the techniques incorporated in the \ac{NN} structure.
	Lastly, \ac{FT} outperforms TC-DQN$^+$ in two cases when the traffic flow is randomly generated in every direction (scenario 3).
	This is partly due to the form of reward function defined in \Cref{eq:reward} as for complex environments, the $r_a$ and $r_e$ may not be clearly translated into an interpretable measure to the agent (for example, traffic scenario 3 of case 3).
	We foresee that refining the reward function with more theoretical grounding could resolve this shortcoming.
	This is addressed as a future direction.
	
	\textit{3) Ablation Plots.}	As mentioned in \Cref{subsec:neuralNet}, our methodology is a combination of several recent techniques from \ac{RL} community.
	Accordingly, it is enlightening to realize the potential contribution of each of these methods to the final performance.
	The comparison is summarized in \Cref{fig:example}  to showcase the median of total wait time among ablated variants (results are sampled for normal traffic at case 2).
	The most remarkable improvements are due to \emph{Distributional \ac{RL}} and \emph{Double Q-Learning} where the absence of each technique results in a performance close to \ac{DQN}.
	In addition, \ac{PER} has a considerable role in upgrading the performance supporting effective use of data.
	Moreover, while \emph{NoisyNets} is not contributing to the long-term performance or our algorithm, it enables convergence at fewer than 8000 episodes.
	Lastly, the portion of contributions of these techniques may vary according to the complexity of the environment.
	For instance, in the example that was provided in this section, \emph{Dueling Q-Learning} does not contribute significantly as $|\mathcal{A}|$ is not large ($|\mathcal{A}|=2$).
	This could change for the case of environment 3 where $|\mathcal{A}|=4$ and the advantage stream of the \ac{NN} is more discerned.


\section{CONCLUSIONS}
	\label{sec:conclusion}
	
	In this paper, we study a novel approach to integrate deep \ac{RL} into traffic signal control at different urban intersections.
	We train our model, TC-DQN$^+$, and test it on three different intersections ordered from simplest to most complex and generate traffic scenarios based on real data and on-site observations.
	Our results demonstrate significant improvement in the performance compared with the formerly employed methods using vanilla \ac{DQN}.

	We acknowledge that the current work is only one step towards faster and more reliable intelligent traffic control and can be improved in multitude of directions.
	One potential future route is to further examine the structure of the underlying \ac{NN} to get more accurate nonlinear approximation of the traffic behavior.
	Moreover, there is room to combine our proposal with classic \ac{ATSC} approaches for a better theoretical grounding of planning strategies.
	Last but not least, generalizability is crucial for data-driven traffic schemes.
	Hence, a potential next step is to make TC-DQN$^+$ more robust to unpredictable phenomena such as roadblocks, constructions, accidents, etc.


\bibliographystyle{ieeetr}
\bibliography{citations}


\newpage 

\section{Appendix: Reproducibility Guide}
\label{appendix}

In this section, we provide necessary documentation to make the results of this work reproducible.
We go over the required of hyperparameters and introduce the \ac{RL} environments parameterization from \Cref{sec:experiments}.
Note that the hyperparameters highly depend on the given problem.
The values provided in this section regard test case 3 in our experiments.
The psuedocode of TC-DQN$^+$ algorithm is also provided to facilitate implementation.

\vspace{1mm}

\textit{\ac{RL} Exploration.} Whenever \emph{NoisyNets} is off, we use $\epsilon$-greedy exploration as,
\begin{align*}
    \epsilon_t = \epsilon_f + (\epsilon_i - \epsilon_f) \exp({-t/\epsilon_d}),
\end{align*}
where $\epsilon_t$ is the exploration factor at frame $t$, $\epsilon_i=1$ is the initial value, $\epsilon_f=0.05$ denotes the final exploration residual, and $\epsilon_d=15000$ is the decay factor.
Otherwise, when using \emph{NoisyNets}, we set $\epsilon_t=0$ and adjust $\sigma_0=0.4$ to initialize the weights of the noisy layers (see \cite{fortunato2018noisy} for more details).
The discount factor and episode length of the algorithm are set to $\gamma=0.99$ and $T=400s$ respectively.

\textit{\ac{NN} Parameters.} The \ac{NN} takes data mini-batches of size $m=32$.
The input, output, linear (fully-connected), and noisy layers each contains $n=41$, $|\mathcal{A}|=4$, $n_{fc}=512$, and $n_{nl}=64$ neurons respectively.
The target network is updated with the frequency $t_f=10,000$ frames and we use Adam optimization \cite{kingma2014adam} with learning rate $\xi=0.0002$.

\textit{\ac{PER}}. We consider a buffer memory size of $M=2^{20}$ samples.
The priorities of transitions along with the importance sampling weights defined as,
\begin{align*}
    p_i \propto \Big| \text{TD}_{\textit{target}} - Q(s,a;\theta) \Big|^{\omega_i}, \hspace{9mm}
    \omega_i = \big( \frac{1}{M} \frac{1}{P(i)} \big)^{\beta}, \hspace{9mm} P(i)=p_i^\alpha / \sum_k p_k^\alpha,
\end{align*}
where $P(i)$ is the probability of transition $i$, $\beta$ is the annealing factor initialized from $\beta_0=0.4$ and linearly increased to 1 with increments of $\Delta \beta = 0.001$, and $\alpha=0.6$ adjusts the prioritization.
The priorities are offset by $\varepsilon=0.01$ to prevent data from not being revisited when the TD-error is close to zero (see \cite{Schaul2016PrioritizedER} for further details).

\textit{Distributional RL.} Following the machinery in \cite{bellemare2017distributional}, we use discrete support $z$ to approximate the distribution of returns instead of the expected return defined by,
\begin{align*}
    z^i=\nu_{\min} + (i-1) \frac{\nu_{\max} - \nu_{\min}} {N_{\text{dist}}-1}, \hspace{8mm} \text{for} \hspace{2mm} i\in\{1,\dots,N_{\textbf{dist}}\},
\end{align*}
where $\nu_{\min}=-4$ and $\nu_{\max}=4$ are the minimum and maximum values of the distribution and $N_{\textbf{dist}}=41$ denotes the number of discrete bins (atoms).

\textit{Environment Settings.} We set $T_g=10s$, $T_y=3s$, and $T_r=2s$ for the green, yellow, and all-red phase intervals respectively.
For the episodic reward (defined in \eqref{eq:re}), we set the parameters $a=3.5$, $b=-0.5$, $\eta=0.007$, and $\zeta=1000$.
For the action-based reward, we let the weights $p_1=0.002$, $p_2=0.01$, and $p_3=0.1$.
The vehicles' speed limit is set equal to $v=40\  \text{km}/\text{h}$ and sensors range is set to $d=40m$.

\textit{Traffic Parameters.} The three environment cases (see \Cref{fig:caseStudies}) are different in the number of actions and complexity of the states (due to traffic level, lane changes, etc.).
In particular, in case 1 (simplest case) left turns are not allowed (for instance $S\rightarrow W$, right turns are permitted in relevant lanes).
In case 2, left turns are allowed but do not pose new actions (no separate light phase defined for left turns).
Therefore, the crossing vehicles have to wait due to right of way of vehicles from the opposite direction.
This is one of the main complexity factors in the test case scenarios with higher traffic intensity.
Finally, Case 3 is the most complex environment where all types of signal phases are allowed.

\Cref{tab:tabEnvs} provides the parameters we used in the simulation to create these environments.
The traffic level parameters are the probability of vehicle generation on corresponding lanes and due to different traffic levels as defined in \ac{SUMO}.
To find these probabilities, we use data from \cite{openData} and also conducted on-site in-person observations at intersections given in \Cref{fig:caseStudies} in San Francisco Bay Area.
This was an effort to make the input to the simulations more realistic as \ac{RL} trial-and-error is high-risk and impractical on real urban traffic.

\begin{table}[H]
    \centering
    \includegraphics[width=0.6\columnwidth]{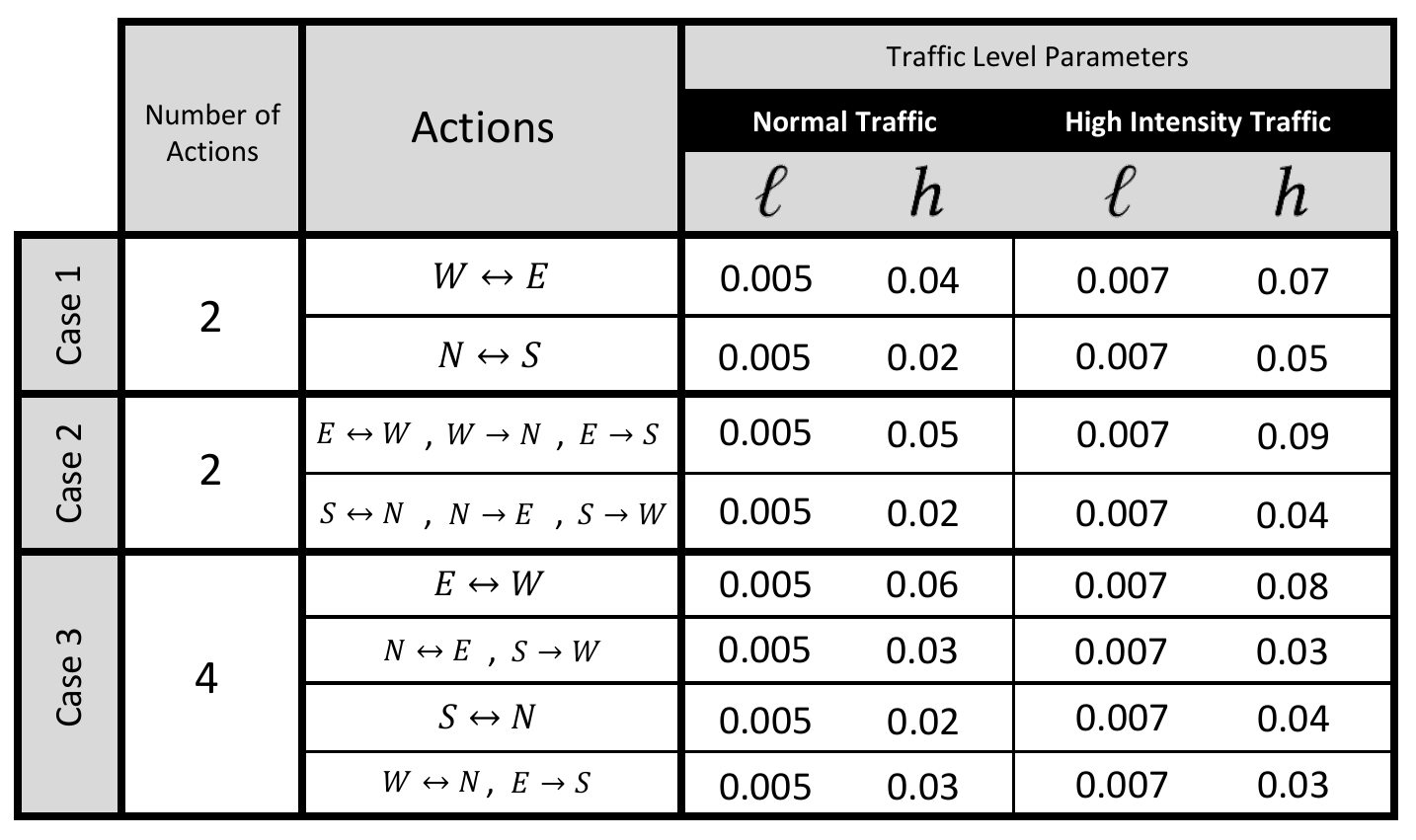}
    \caption{Traffic settings of our experiments.}
    \label{tab:tabEnvs}
\end{table}

\noindent \textbf{Source Code Instructions.} For the implementation of this work, we have used \textit{python} 3.7.4, \textit{PyTorch} 1.2 \cite{NEURIPS2019_9015}, \textit{OpenAI gym} \cite{brockman2016openai}, and \ac{SUMO} 1.3.1 \cite{krajzewicz2002sumo}.
The codes of our experiments are accessible on the \href{https://github.com/siavash91/TC-DQN}{GitHub repository} of this project.

\vspace{20mm}

\newpage

\begin{algorithm}[H]
	\label{alg:1}
	\caption{TC-DQN$^+$ Algorithm}
	\begin{algorithmic}[1]
	    \State \textbf{Initialize}: Network with $\theta$ and target network with $\theta^-=\theta$
		\State \textbf{Input}: replay buffer $\mathcal{D}$ with memory $M$, random variables $\varepsilon$ for the noisy networks, mini-batch size $m$, learning rate $\eta$, target update frequency $t_f$, discount factor $\gamma$, length of episodes $T$, hyperparameters for distributional \ac{RL} $N_{\text{dist}}$, $\nu_{\min}$, $\nu_{\max}$, and hyperparameters for \ac{PER} $\alpha$, $\beta$
		
		\vspace{1em}
		
		\State \textbf{For} episode $e\in\{1,2,\dots\}$ \textbf{do}
		\State $\hspace{6mm}$ \textbf{For} $t=\{1,\dots,T\}$ \textbf{do}
		\State $\hspace{12mm}$ Initialize state $s_0\sim$ \emph{env}
		\State $\hspace{12mm}$ Choose action $a_t\leftarrow\text{argmax}_{a'\in\mathcal{A}}Q(s_t,a',\theta)$
		\State $\hspace{12mm}$ Run the environment and Sample $s_{t+1}$ 
		\State $\hspace{12mm}$ Obtain $r_t = r_{a_t} + r_e$ from \eqref{eq:reward}, \eqref{eq:ra}, and \eqref{eq:re}
		\State $\hspace{12mm}$ Store transition $(s_t, a_t, r_t, s_{t+1})$ in the buffer
		\State $\hspace{12mm}$ \textbf{If} $|\mathcal{D}|>M$ \textbf{do}
		\State $\hspace{18mm}$ Delete oldest transition from $\mathcal{D}$
		\State $\hspace{12mm}$ Sample mini-batch of size $\min\{m, |\mathcal{D}|\}$ from $\mathcal{D}$
		\State $\hspace{12mm}$ Sample noisy variables for: \hspace{3mm} \textcolor{cyan}{\texttt{\% Noisy networks initialization}}
		\State $\hspace{18mm}$ the online network $\lambda\sim\varepsilon$
		\State $\hspace{18mm}$ the target network $\bar{\lambda}\sim\varepsilon$
		\State $\hspace{18mm}$ the action selection network $\bar{\bar{\lambda}}\sim\varepsilon$
		\State $\hspace{12mm}$ \textbf{For} $i\in\{1,\dots,m\}$ \textbf{do} \hspace{3mm} \textcolor{cyan}{\texttt{\% Sampling from prioritized replay buffer}}
		\State $\hspace{18mm}$ Sample transition $i\sim P(i)=p_i^{\alpha}/\sum_j p_j^{\alpha}$
		\State $\hspace{18mm}$ Find PER weight $\omega_i=(MP(i))^{-\beta}/\max_j\omega_j$
		\State $\hspace{18mm}$ Set $\text{TD}_{\text{target}}=r_i+\gamma \hat{Q}(s_i,\text{argmax}_{a'}Q(s_i,a',\theta,\lambda),\theta^-,\bar{\lambda})$
		\State $\hspace{18mm}$ Find TD-error $\delta_i = \text{TD}_{\text{target}} - Q(s_{i-1},a_{i-1}, \theta, \lambda)$
		\State $\hspace{18mm}$ Update transition priority $p_i\leftarrow |\delta_i|$
		\State $\hspace{18mm}$ \textbf{If} $s_{t+1}$ is a terminal state \textbf{do}
		\State $\hspace{26mm}$ $\hat{Q}\leftarrow r_i$
		\State $\hspace{18mm}$ \textbf{Else do}
		\State $\hspace{26mm}$
		$\hat{Q}(s_i,b',\theta^-,\bar{\lambda})\leftarrow \sum_{j}z_jp_j(s_i,b')$
		\State $\hspace{26mm}$ $b\leftarrow\text{argmax}_{b'\in\mathcal{A}}\hat{Q}(s_i,b',\theta^-,\bar{\bar{\lambda}})$ \hspace{3mm} \textcolor{cyan}{\texttt{\% action with double Q-learning}}
		\State $\hspace{26mm}$ $m_j\leftarrow 0$ for $j=\{0,\dots,N_{\text{dist}}-1\}$
		\State $\hspace{26mm}$ \textbf{For} $k=\{1,\dots,N_{\text{dist}}\}$ \textbf{do} \hspace{3mm} \textcolor{cyan}{\texttt{\% Discrete distribution of returns}}
		\State $\hspace{32mm}$ $Tz_k\leftarrow[r_t+\gamma z_k]_{\nu_{\min}}^{\nu_{\max}}$
		\State $\hspace{32mm}$ $c_k\leftarrow (Tz_k-\nu_{\min})/\Delta z$
		\State $\hspace{32mm}$ $l\leftarrow\floor*{c_k},\ u\leftarrow\ceil*{c_k}$
		\State $\hspace{32mm}$ $m_l\leftarrow m_l+p_k(s_{t+1},b_t)(u-c_k)$
		\State $\hspace{32mm}$ $m_u\leftarrow m_u+p_k(s_{t+1},b_t)(c_k-l)$
		\State $\hspace{26mm}$ Set Loss $\mathcal{L} = -\sum_j m_j \omega_i \log p_j(s_t,a_t)$
		\State $\hspace{26mm}$ Do a gradient step with $\theta\leftarrow\theta-\eta\nabla\mathcal{L}$
		\State $\hspace{6mm}$ \textbf{If} $e\times t\equiv 0$ (mod $t_f$) \textbf{do}
		\State $\hspace{12mm}$ Update the target network $\theta^-\leftarrow\theta$
	\end{algorithmic}
\end{algorithm}

\end{document}